\documentclass[a4paper, 10pt, journal]{article}
  
\usepackage[colorlinks,%
            bookmarksopen,%
            bookmarksnumbered,%
            citecolor=red,%
            urlcolor=red]{hyperref}

\usepackage[utf8]{inputenc}
\usepackage[T1]{fontenc}

\usepackage[letterpaper, margin=2cm]{geometry}

\usepackage{array,multirow,makecell}
\newcolumntype{R}[1]{>{\raggedleft\arraybackslash }b{#1}}
\newcolumntype{L}[1]{>{\raggedright\arraybackslash }b{#1}}
\newcolumntype{C}[1]{>{\centering\arraybackslash }b{#1}}
\newcolumntype{M}[1]{>{\centering\arraybackslash}m{#1}}

\usepackage{caption}

\usepackage{graphicx}
\usepackage{mathptmx}
\usepackage{amssymb}
\usepackage{amsmath} 
\usepackage{amsfonts}
\usepackage{booktabs}    
\usepackage[squaren,binary,Gray]{SIunits} 

\usepackage[inline]{enumitem}
\setlist[itemize]{topsep=0pt,itemsep=-1ex,partopsep=1ex,parsep=1ex,label=$\centerdot$}
    
\begin{document}

\title{Paris-Lille-3D: a large and high-quality ground truth urban point cloud
dataset for automatic segmentation and classification}

\author{\hspace{-1.5cm} Xavier Roynard, Jean-Emmanuel Deschaud and François Goulette\\ 
 \\
 \hspace{-1.0cm}\{xavier.roynard ; jean-emmanuel.deschaud ; francois.goulette\}@mines-paristech.fr
 \\ 
 \\
\hspace{-1.2cm}Mines ParisTech, PSL Research University, Centre for Robotics}

\date{}

\maketitle

\begin{abstract}
This paper introduces a new Urban Point Cloud Dataset for Automatic Segmentation and Classification acquired by Mobile Laser Scanning (MLS). We describe how the dataset is obtained from acquisition to post-processing and labeling. This dataset can be used to train pointwise classification algorithms, however, given that a great attention has been paid to the split between the different objects, this dataset can also be used to train the detection and segmentation of objects.
The dataset consists of around $2\kilo\meter$ of MLS point cloud acquired in two cities. The number of points and range of classes make us consider that it can be used to train Deep-Learning methods.
Besides we show some results of automatic segmentation and classification. 
The dataset is available at: \href{http://caor-mines-paristech.fr/fr/paris-lille-3d-dataset/}{http://caor-mines-paristech.fr/fr/paris-lille-3d-dataset/}.
\end{abstract}

\begin{center}
\end{center}

\begin{figure}\centering
 \includegraphics[width=\linewidth]{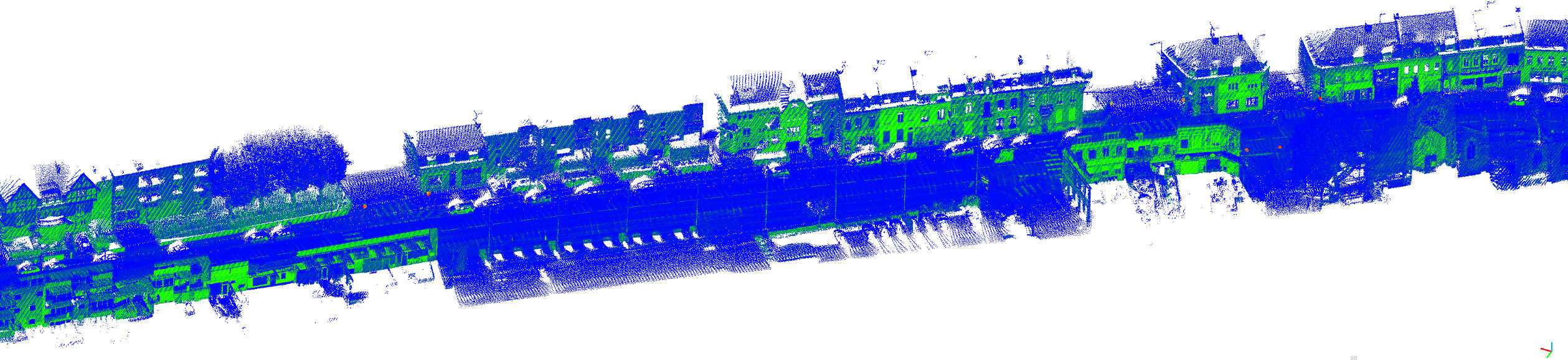}\\
 \includegraphics[width=\linewidth]{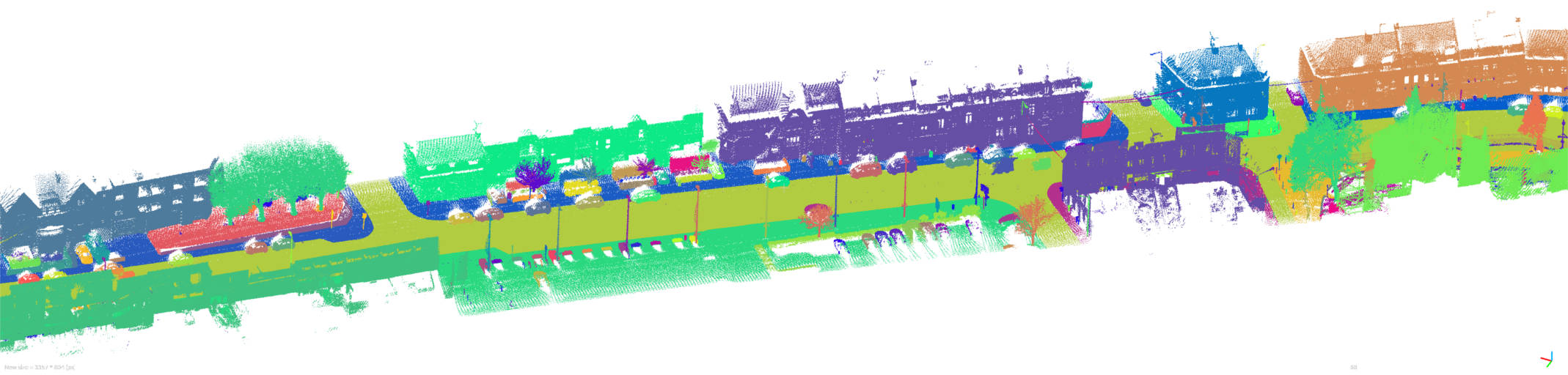}\\
 \includegraphics[width=\linewidth]{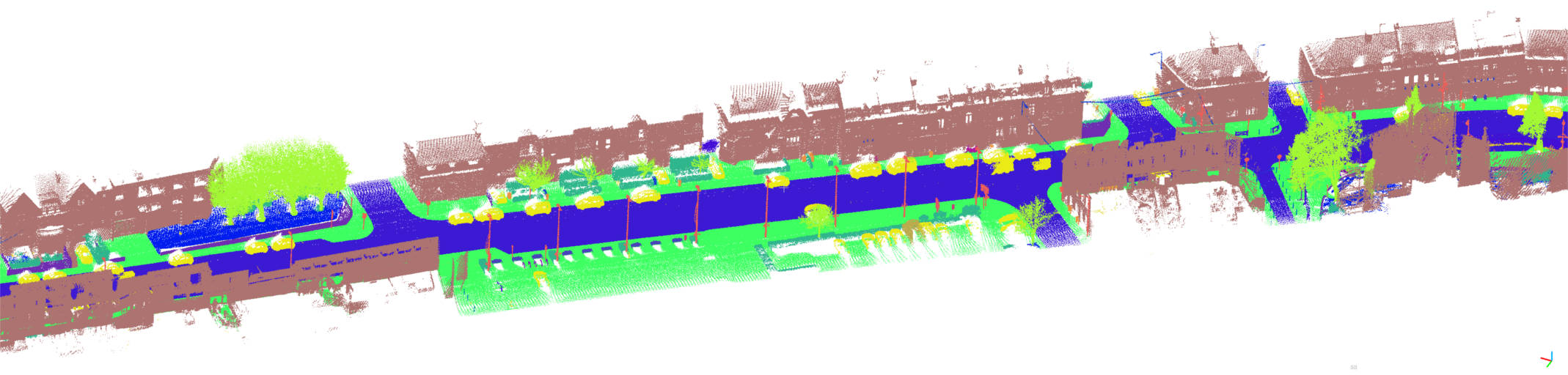}
 \caption{Part of our dataset. Top: reflectance from blue(0) to red(255), middle: object label (different color for each object), bottom: object  class (different color for each class). \label{fig:dataset}}
\end{figure}


\section{Introduction}

With the development of segmentation and classification methods of 3D point clouds by machine-learning, more and more data are needed in quantity and quality (number of points, number of classes, quality of segmentation).

There are always more datasets of classification and segmentation of images, visual and LiDAR odometry or SLAM, detection of vehicles and pedestrians on videos, stereo-vision, optical flow, etc. But it is still difficult to find datasets of segmented and classified urban 3D point clouds. The only comparable datasets are the ones described in section \textit{Available Datasets}. Each of them has its advantages and disadvantages, but we believe that none of them combines all the advantages of the dataset we publish.

In section \textit{Our Dataset: Paris-Lille-3D}, we present a new urban dataset that we have created, where the objects are sufficiently segmented that the task of segmentation can be learned very precisely.
Our dataset can be found at the following address: \href{http://caor-mines-paristech.fr/fr/paris-lille-3d-dataset/}{http://caor-mines-paristech.fr/fr/paris-lille-3d-dataset/}. 

In section \textit{Results of automatic segmentation and classification}, we give some results of automatic segmentation and classification on our dataset.

\section{Available Datasets}
\label{sec:otherDatasets}

Numerous datasets are used for training and benchmarking machine learning algorithms. The provided data allows for the training of methods that perform a given task, and the evaluation of performance on a test set allows evaluation of the quality of the results obtained and comparison of different methods according to various metrics.

Many different tasks can be learned, the most common being classification (for example, for an image, it is giving the class of the principal object visible). Another task may be to segment the data into its relevant parts (for the images it is grouping all the pixels that belong to the same object). There are multiple other tasks that can be learned, from image analysis to translation in natural language processing, for a survey see \cite{ferraro2015survey}.


There is a bunch of existing datasets in many fields. Each dataset has different types of data, in type (image, sound, text, point clouds, graphs), quantity (from hundreds to billion of samples), quality, number of classes (from tens to thousands), and tasks to learn. Amongst the most famous are:
\begin{itemize}
 \item image classification and segmentation datasets:  \href{http://www.image-net.org/}{ImageNet} 
  \cite{deng2009imagenet}, \href{http://mscoco.org/}{MS COCO} 
  \cite{lin2014microsoft},
 \item stereovision dataset for depth map estimation: \href{http://vision.middlebury.edu/stereo/data/}{Middlebury Stereo Datasets} 
  \cite{scharstein2014high},
 \item video dataset: \href{https://research.google.com/youtube8m/}{Youtube-8M} 
  \cite{abu2016youtube},
 \item odometry, stereovision, optical flow and 3D object detection dataset: \href{http://www.cvlibs.net/datasets/kitti/index.php}{KITTI} 
  \cite{Geiger2012CVPR},
 \item SLAM dataset: \href{http://robots.engin.umich.edu/SoftwareData/Ford}{Ford Campus Vision and Lidar Data Set} \cite{pandey2011ford},
 \item long-term localization datasets: \href{http://robotcar-dataset.robots.ox.ac.uk/}{the Oxford Robotcar Dataset} \cite{RobotCarDatasetIJRR} and \href{http://robots.engin.umich.edu/nclt/}{the NCLT Dataset} \cite{carlevaris2016university},
 \item urban street image segmentation dataset: \href{https://www.cityscapes-dataset.com/}{The Cityscapes Dataset} \cite{cordts2016cityscapes}.
\end{itemize} 
Closer to our field of research are Airborne Laser Scanning (ALS) datasets as provided with the \href{http://www2.isprs.org/commissions/comm3/wg4/3d-semantic-labeling.html}{3D Semantic Labeling Contest} 
  \cite{niemeyer2014contextual}.

The data we are interested in are urban 3D point clouds. There are mainly two methods that allow to acquire these data in quality sufficient for us:
\begin{itemize}
 \item Mobile Laser Scanning (MLS), with a LiDAR mounted on a ground vehicle or a drone. To register the clouds, an accurate 6D-pose of the vehicle must be known.
 \item  Terrestrial Laser Scanning (TLS) by static LiDAR, the LiDAR must be moved between each acquisition and clouds must be registered.
\end{itemize}
The ALS does not allow to obtain a sufficient density of points because of the distance and the angle of acquisition.

There are already some segmented and classified urban 3D point cloud datasets. However these datasets are very heterogeneous and each has features that can be seen as defects. In the 4 next sub-sections, we make a comparison of the existing datasets and identify their strengths and weaknesses for automatic classification and segmentation. Table \ref{tab:comparison} presents a quantitative comparison of these datasets with ours.

\begin{table*}\centering
 \begin{tabular}{C{2.3cm}C{2.5cm}C{2.0cm}C{2.2cm}C{2.2cm}}
  \toprule
   \textbf{Name} & \textbf{Lidar type} & \textbf{Length} & \textbf{Number of points} & \textbf{Number of classes} \\\midrule
   Oakland & MLS mono-fiber & $1510\meter$ & $1.61\mega$ & $44$ \\\midrule
   Semantic3D & static LiDAR & -- & $1660\mega$ & $8$ \\\midrule
   Paris-rue-Madame & MLS multi-fiber & $160\meter$ & $20\mega$ & $17$ \\\midrule
   IQmulus & MLS mono-fiber & $210\meter$ & $12\mega$ & $22$ \\\midrule
   Paris-Lille-3D & MLS multi-fiber & $1940\meter$ & $143.1\mega$ & $50$ \\
  \bottomrule 
 \end{tabular}
 \captionof{table}{Comparison of urban 3D point cloud datasets.\label{tab:comparison}}
\end{table*}

\subsection{\href{http://www.cs.cmu.edu/~vmr/datasets/oakland_3d/cvpr09/doc/}{Oakland 3-D Point Cloud Dataset} \cite{munoz2009contextual}}\label{subsec:oakland}
This dataset was acquiered by a MLS system mounted with a side looking Sick monofiber LiDAR. Since it is a mono-fiber LiDAR, it has the disadvantage of hitting the objects from a single point of view, so there are many occlusions. In addition it is much less dense than other datasets because of the low acquisition rate of the LiDAR (see figure \ref{fig:oakland}).
In addition, this dataset contains 44 classes of which a large part (24) have less than 1000 labeled points. This is very low to be able to distinguish objects, especially when these points are distributed over several samples.

\begin{center}\centering
 \includegraphics[width=\linewidth]{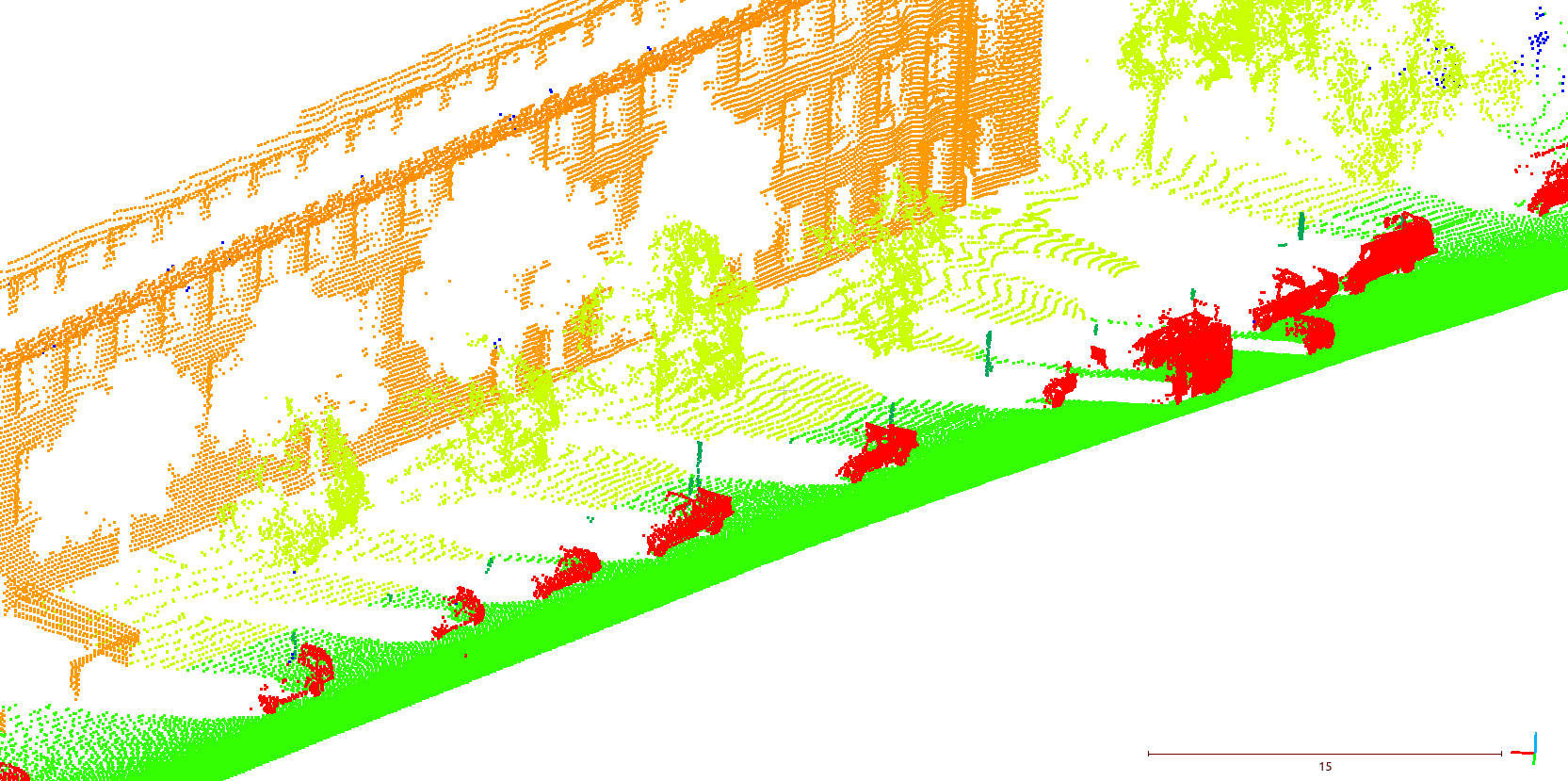}
 \captionof{figure}{Example of cloud in Oakland dataset. Low density, few classes, big shadows behind trees (due to monofiber LiDAR).\label{fig:oakland}}
\end{center}

\subsection{\href{http://www.semantic3d.net/}{Semantic3D} \cite{hackel2016fast}}\label{subsec:semantic3d}
This dataset was acquired by static laser scanners. It is therefore much more precise and dense than a dataset acquired by MLS, but it has disadvantages inherent to static LiDARs (see figure \ref{fig:semantic3d}):
\begin{itemize}
 \item The density of points varies considerably depending on the distance to the sensor.
 \item There are occlusions due to the fact that sensors do not turn around the objects. Even by registering several clouds acquired from different viewpoints, there are still a lot of occlusions.
 \item The acquisition time is much more important than by MLS, which prevents to obtain very miscellaneous scenes.
\end{itemize}

\begin{center}
 \includegraphics[width=\linewidth]{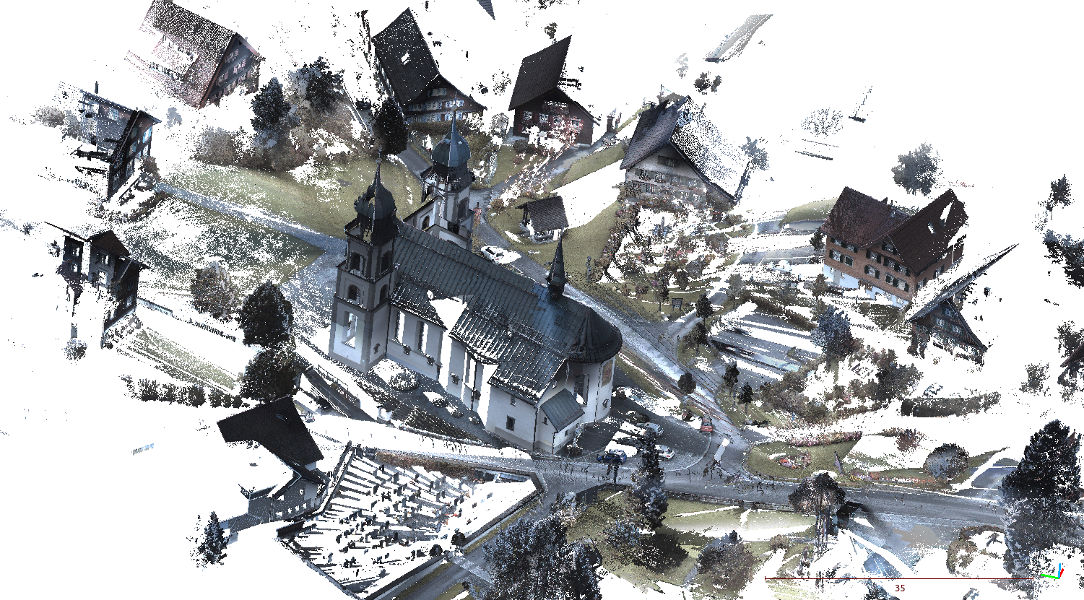}
 \captionof{figure}{Example of cloud in Semantic3D dataset (3 clouds registered). Occlusions, density depends on the distance to the LiDAR.\label{fig:semantic3d}}
\end{center}

\subsection{\href{http://cmm.ensmp.fr/~serna/rueMadameDataset.html}{Paris-rue-Madame Database} \cite{serna2014paris}}\label{subsec:ruemadame}
This dataset was acquired by an earlier version of our MLS system \cite{goulette2006integrated}. 
This dataset was segmented and annotated semi-automatically, first by a mathematical morphology method on elevation images \cite{serna2014paris} and then refined by hand. Some segmentation inaccuracies at the edges of objects remain (see figure \ref{fig:ruemadame}), in particular the bottom of the objects is annotated as belonging to the ground.
Moreover, the system as well as the point cloud processing pipeline have been greatly improved. We can now generate clouds much less noisy.

\begin{center}\centering
 \includegraphics[width=\linewidth]{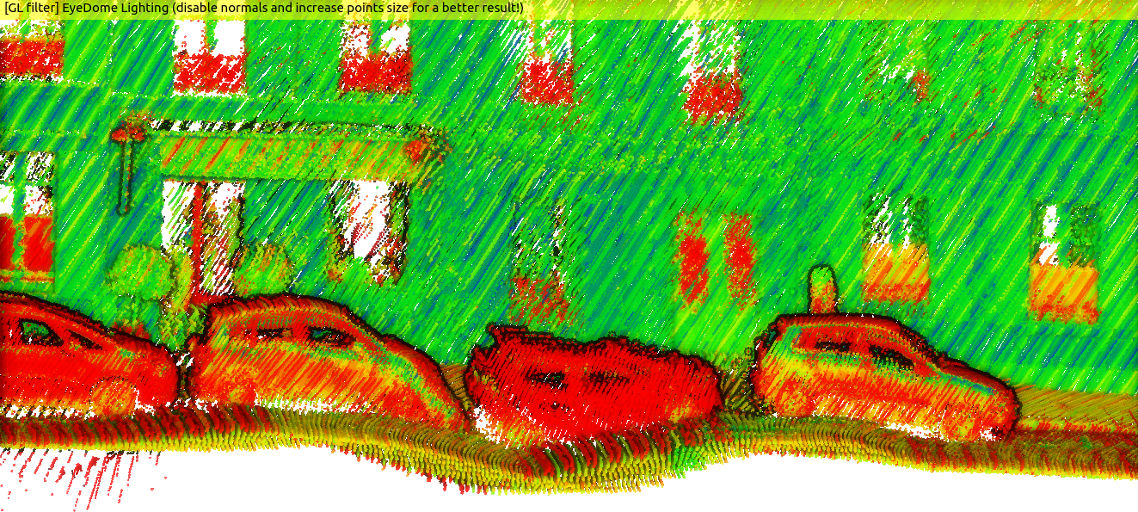}\vspace{0.1cm}
 \includegraphics[width=\linewidth]{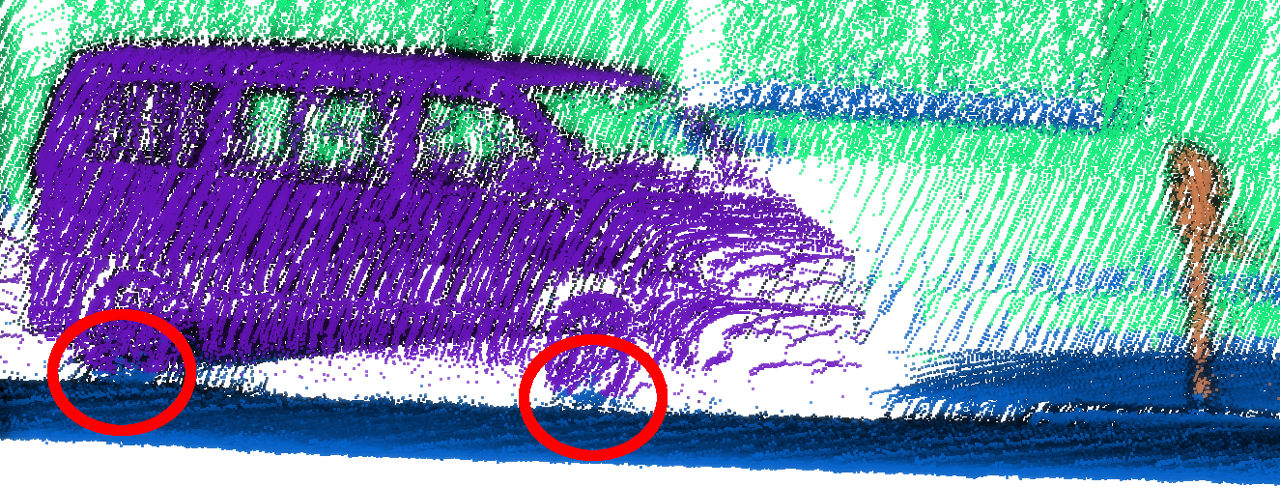}
 \captionof{figure}{Example of cloud in Rue-Madame dataset. Ground truth mistakes: can be very noisy (top), parts of cars are seen as road (bottom).\label{fig:ruemadame}}
\end{center}

\subsection{\href{http://data.ign.fr/benchmarks/UrbanAnalysis/}{IQmulus \& TerraMobilita Contest} \cite{vallet2015terramobilita}}\label{subsec:iqmulus}
This dataset was acquired by a MLS system mounted with a monofiber \href{http://www.riegl.com/nc/products/industrial-scanning/produktdetail/product/scanner/10/}{Riegl LMS-Q120i} LiDAR. This LiDAR has the advantage of being more accurate than a multi-fiber LiDAR such as the \href{http://velodynelidar.com/hdl-32e.html}{Velodyne HDL-32E}, but it is also more expensive. Moreover, since it is mono-fiber, it has the disadvantage of hitting the objects from a single point of view, so there are many occlusions.

For the annotation, the scan lines of the LiDAR were concatenated one above the other to form 2D images. The values of the pixels are the intensity of laser return.

This method has the advantage of being easy to put into production, which allowed the IGN to annotate a large dataset. However, inaccuracies in countouring annotation of 2D images 
generate badly classified points, the points around the occlusions are classified in the class of the object that creates the occlusion.
(see figure \ref{fig:terramobilita}).

\begin{center}\centering
 \includegraphics[width=\linewidth]{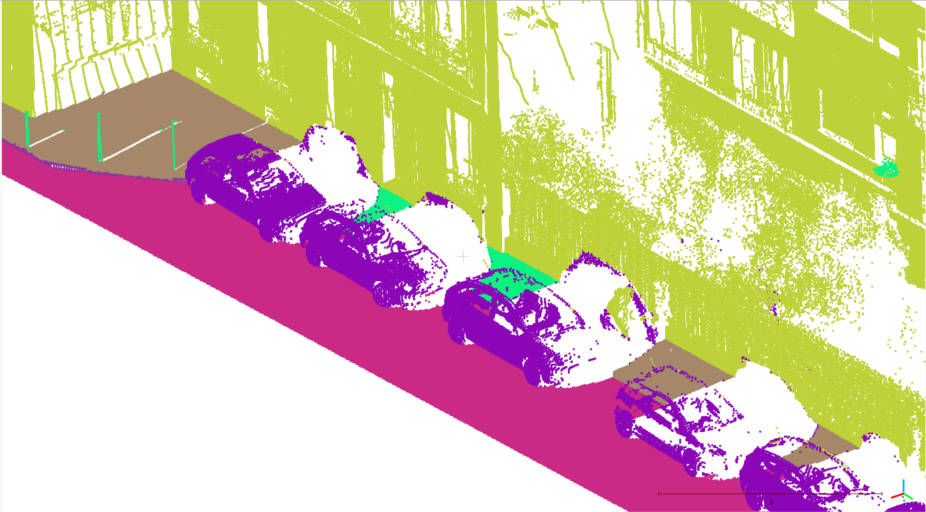}
 \captionof{figure}{Example of cloud in iQmulus/TerraMobilita dataset. As a monofiber LiDAR is used, there are shadows behind objects. Moreover points of the wall behind cars are classified as car.\label{fig:terramobilita}}
\end{center}

\section{Our Dataset: Paris-Lille-3D} \label{sec:ourDataset}

\subsection{Acquisition} \label{subsec:acquisition}
All point clouds used in our dataset were acquired with the MLS prototype of the center for robotics of Mines ParisTech: L3D2 \cite{goulette2006integrated} (as seen in figure \ref{fig:L3D2}). It is a Citro\"en Jumper equipped with a GPS (Novatel FlexPak 6), an IMU (Ixsea PHINS in LANDINS mode) and a Velodyne HDL-32E LiDAR mounted at the rear of the truck with an angle of 30 degrees between the axis of rotation and the horizontal.

\begin{center}\centering
 \includegraphics[trim=0 0 0 0,clip,width=\linewidth]{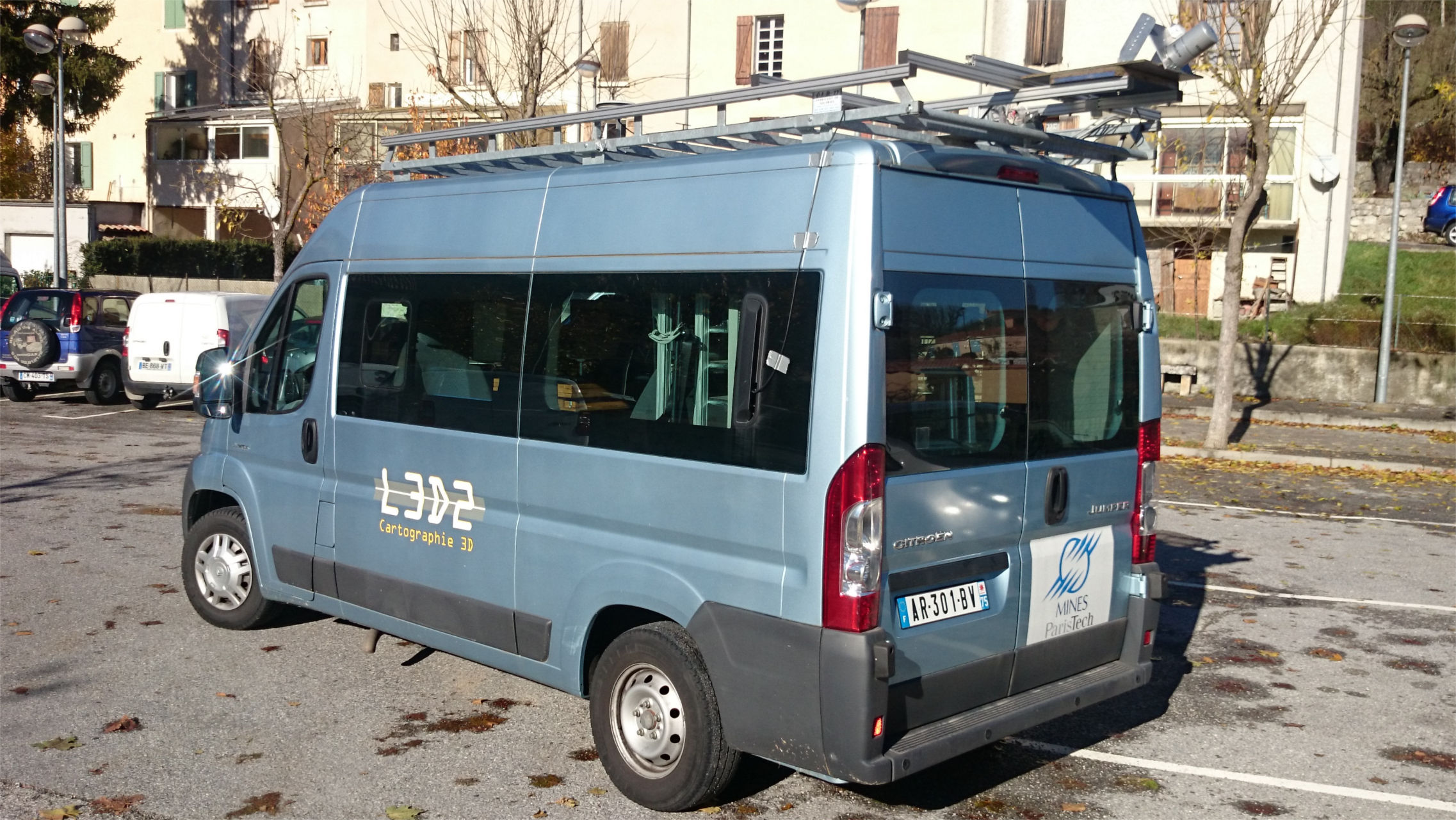}
 \captionof{figure}{MLS prototype: L3D2\label{fig:L3D2}}
\end{center}

For localization, we use a dual-phase L1/L2 RTK-GPS at $1 \hertz$ with a fixed base provided by the IGN \href{http://rgp.ign.fr/}{RGP}\footnote{\url{http://rgp.ign.fr/}} (Permanent GNSS Network). RGP bases are: SMNE for Paris dataset and LMCU for Lille dataset. The IMU sends data at $100 \hertz$.
Data from the LiDAR and IMU are synchronised thanks to  PPS signal from the GPS.

In post-process, we retrieve data from RGP fixed base, and we generate the trajectory with the \href{https://www.novatel.com/products/software/inertial-explorer/}{Inertial Explorer}\footnote{\url{https://www.novatel.com/products/software/inertial-explorer/}} software. The method used is Tightly Coupled GPS-RTK/INS Kalman Smoothing EKF. We obtain a trajectory in WGS84 system at $100 \hertz$, that we convert to Lambert RGF93.

Then, as each point has its own timestamp, we linearly interpolate the trajectory. Moreover we only keep points measured at a distance less than $20 \meter$ in order to keep only areas of sufficiently high density. Finally we build clouds for which each point is characterized by a vector $(x,y,z,x_{origin},y_{origin},z_{origin},t,i)$, where $i$ is the intensity of the LiDAR return.

We do not apply any method of SLAM, cloud registration or loop closure. All trajectories are built with Inertial Explorer.

\subsection{Description of point clouds} \label{subsec:description_pc}

The dataset consists of three parts, two parts in the agglomeration of Lille and one in Paris (see figure \ref{fig:traj}).

\begin{center}\centering
 \includegraphics[width=\linewidth]{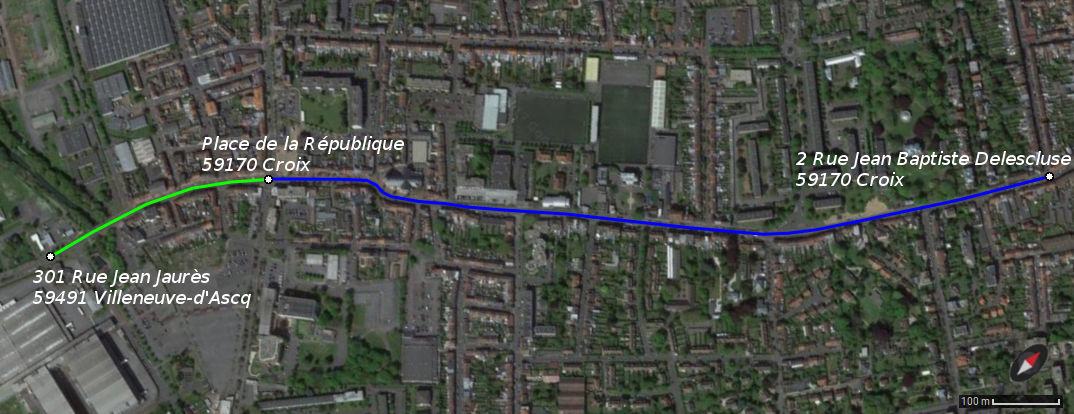}
\\ \vspace{0.1cm}
 \includegraphics[width=\linewidth]{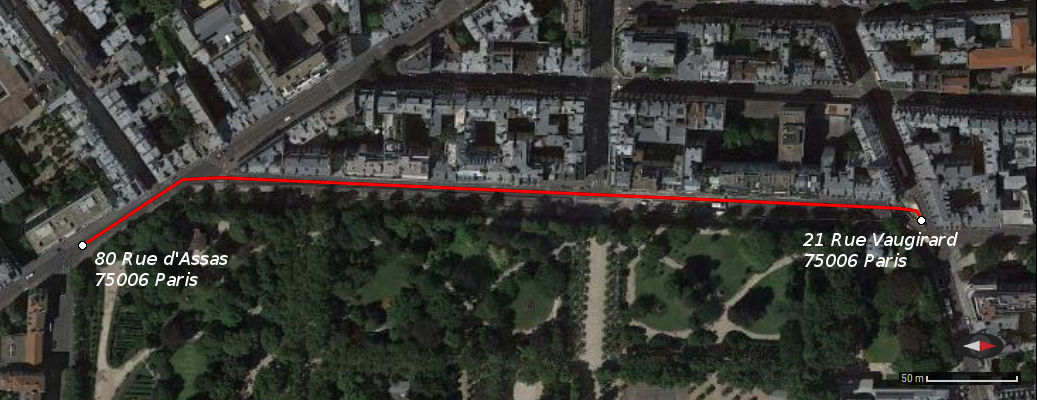}
 \captionof{figure}{Trajectories of the experimental vehicle during acquisition. Top: the 2 trajectories in agglomeration of Lille are in green and blue. Bottom: the trajectory in Paris is in red. (Pictures from Google Maps) \label{fig:traj}}
\end{center}

For the sake of precision, an offset has been substracted in the plane (x,y) to all the points so that they hold in \texttt{float} (32 bits). Data are distributed as explained in table \ref{tab:parts}.
\begin{small}
\begin{table}[h]\centering
 \begin{tabular}{C{1.0cm}C{1.0cm}C{1.3cm}C{3.3cm}}
  \toprule
   \textbf{Section} & \textbf{Length} & \textbf{Number of points} & \textbf{RGF93 Offset} \\\midrule
   Lille1 & $1150\meter$ & $71.3\mega$ & $(711164.0\meter, 7064875.0\meter)$ \\
   Lille2 & $340\meter$ & $26.8\mega$ & $(711164.0\meter, 7064875.0\meter)$  \\
   Paris & $450\meter$ & $45.7\mega$ & $(650976.0\meter, 6861466.0\meter)$  \\\midrule
   \textbf{Total} & $1940\meter$ & $143.1\mega$ & -- \\
  \bottomrule 
 \end{tabular}
 \caption{Description of the three parts of the dataset.\label{tab:parts}}
\end{table}
\end{small}

The clouds have high density with between 1000 and 2000 points per square meter on the ground, but there are some anisotropic patterns due to the multi-beam LiDAR sensor as seen in figure \ref{fig:pattern}.

\begin{center}\centering
 \includegraphics[width=\linewidth]{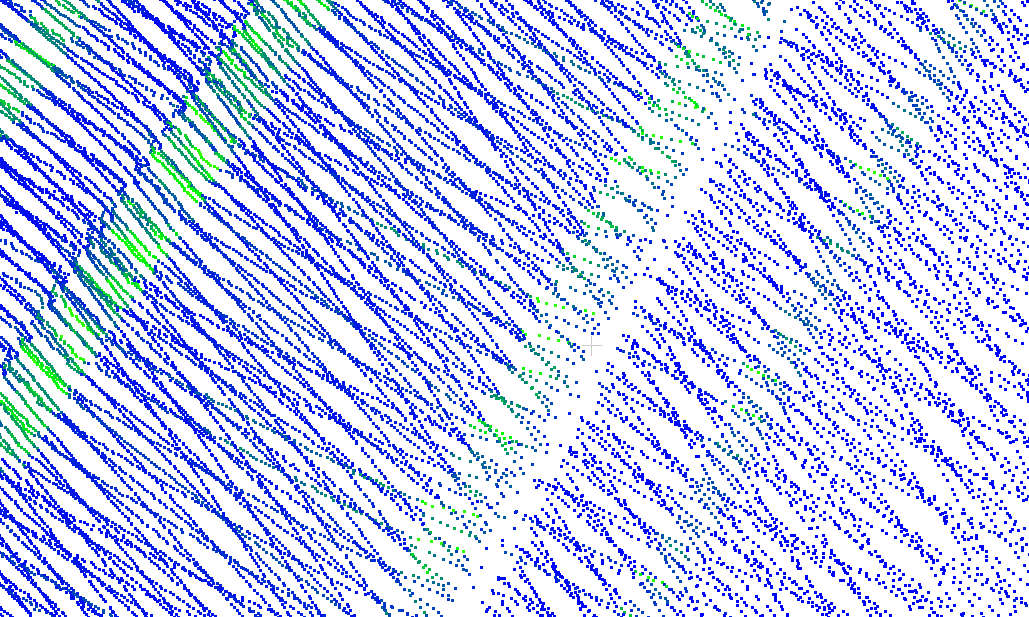}
 \captionof{figure}{Anisotropic pattern on the ground (color of points is the reflectance) \label{fig:pattern}}
\end{center}

\subsection{Description of segmented and classified data} \label{subsec:description_data}

The clouds obtained were segmented and classified by hand using \href{http://www.danielgm.net/cc/}{CloudCompare}\footnote{\url{http://www.danielgm.net/cc/}} software. Some illustrations of the segmented and classified data are shown in figure \ref{fig:dataset}.

We chose to re-use the \href{http://data.ign.fr/benchmarks/UrbanAnalysis/download/classes.xml}{class tree} 
 of iQmulus/Terramobilita benchmark, in which we only change a few classes and add classes relevant to our dataset. It can be found at url: \url{http://data.ign.fr/benchmarks/UrbanAnalysis/download/classes.xml}
For a distribution of number of points by classes, see table \ref{tab:nb_samp}.
Classes added:
\begin{itemize}
 \item \textit{bicycle rack} ($\mathtt{id} = 302021200$)
 \item \textit{statue} ($\mathtt{id} = 302021300$)
 \item \textit{distribution box} ($\mathtt{id} = 302040600$)
 \item \textit{lighting console} ($\mathtt{id} = 302040700$)
 \item \textit{windmill} ($\mathtt{id} = 302040800$)
\end{itemize}
We also change the way vehicles are seen. More precisely, for each class of vehicle, we distinguish sub-classes depending on whether they are parked, stopped (on the road) or moving.
And \textit{Velib terminal} is changed to \textit{bicycle terminal} ($\mathtt{id} = 302021100$) which is more generic.

Except the few classes mentioned above, this class tree appears to be sufficiently complete for classes encountered in our dataset. The XML file describing this tree is named \texttt{classes.xml} and is provided with the dataset. We also provide three ASCII-files (.txt) containing annotations for particular samples. Each line of these files contains: 
\begin{verbatim}
 sample_id, class_id, class_name,
    annotation1, annotation2, ...
\end{verbatim}

The most common annotations are:
\begin{itemize}
 \item "several", for example when trees are interlaced and can not be delimited precisely by hand,
 \item "overturned", for trash cans laid on their side.
\end{itemize}

\begin{center}\centering
 \includegraphics[width=0.050\linewidth, trim={{0.3cm} {-0.7cm} {0.8cm} 0}]{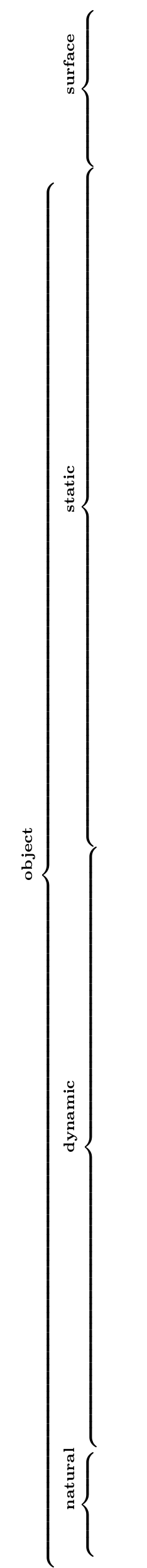}
 \includegraphics[width=0.94\linewidth]{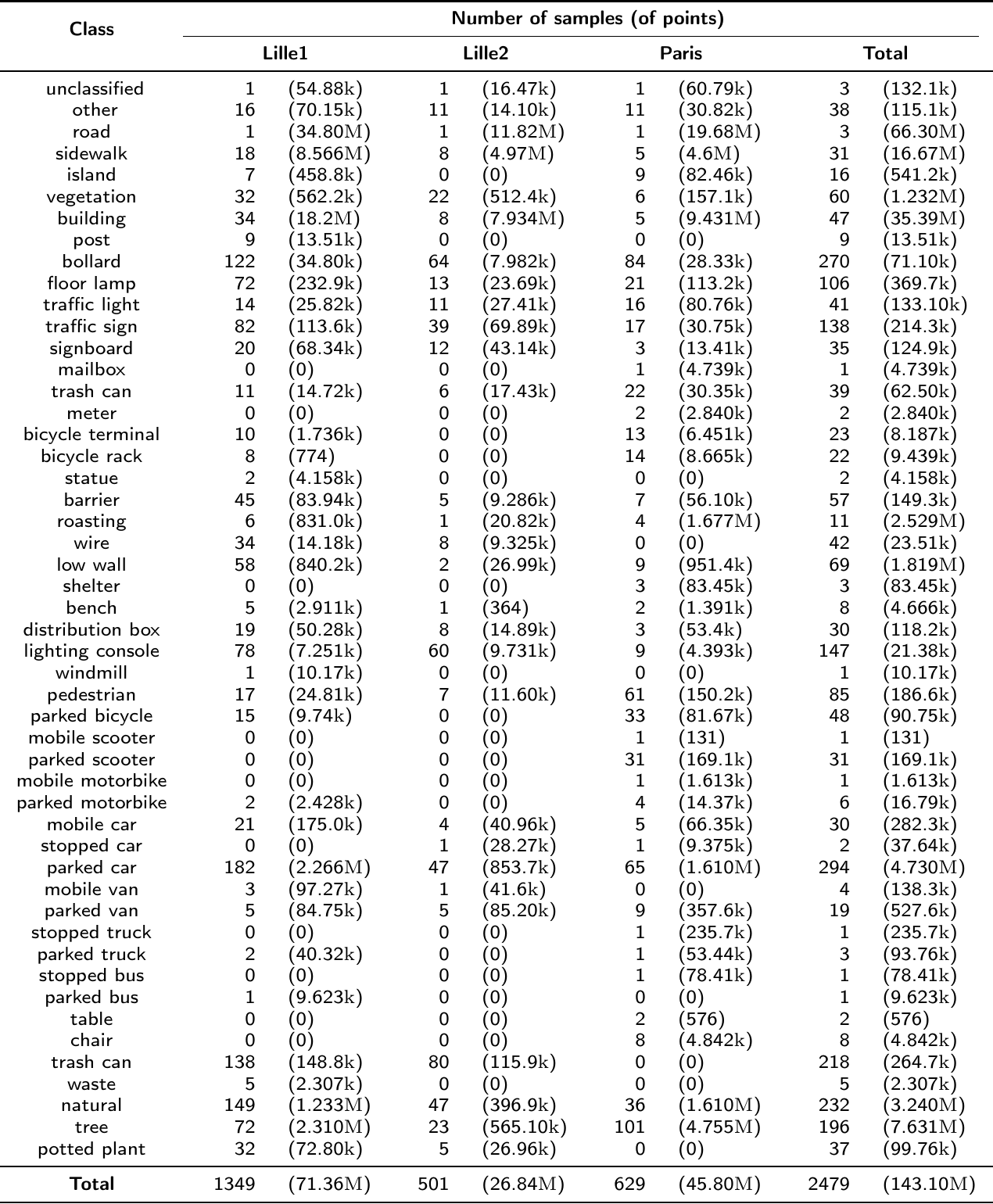}
 \captionof{table}{Number of samples/points for each class ($\kilo$ for thousand and $\mega$ for million). Trash cans appear twice, the first time is for only fixed trash can.\label{tab:nb_samp}}
\end{center}

\subsection{Description of files} \label{subsec:summary}

Each part of the dataset is in a separate PLY-file, a summary of each file can be found in table \ref{tab:overview}. Each point of PLY-files has 10 attributes:
\begin{itemize}
 \item \texttt{x}, \texttt{y}, \texttt{z} (float) : the position of the point,
 \item \texttt{x$\_$origin}, \texttt{y$\_$origin}, \texttt{z$\_$origin} (float) : the position of the LiDAR,
 \item \texttt{GPS$\_$time} (double) : the moment when the point was acquired,
 \item \texttt{reflectance} (uint8) : the intensity of laser return,
 \item \texttt{label} (uint32) : the label of the object to which the point belongs,
 \item \texttt{class} (uint32) : the class of the object to which the point belongs.
\end{itemize}

\begin{center}\centering
 \begin{tabular}{C{1.0cm}C{1.0cm}C{1.4cm}C{1.5cm}C{1.4cm}}
  \toprule
   \textbf{Section} & \textbf{Length} & \textbf{Number of points} & \textbf{Number of objects} & \textbf{Number of classes} \\\midrule
   Lille1 & $1150\meter$ & $71.3\mega$ & $1349$ & $39$ \\
   Lille2 & $340\meter$ & $26.8\mega$ & $501$ & $29$ \\
   Paris & $450\meter$ & $45.7\mega$ & $629$ & $41$ \\\midrule
   \textbf{Total} & $1940\meter$ & $143.1\mega$ & $2479$ & $50$ \\
  \bottomrule 
 \end{tabular}
 \captionof{table}{Overview of our dataset.\label{tab:overview}}
\end{center}

\section{Results of automatic segmentation and classification} \label{sec:ourResults}

In this section, we evaluate an automatic segmentation and classification method on our dataset. There are many approaches to achieve this task, most of them look like one of the following pipelines:
\begin{itemize}
 \item classify each point for example by computing local features (hand-made \cite{weinmann2015semantic} or by Deep-Learning methods \cite{huang2016point}), then group them into objects for example by CRF methods (see \cite{LANDRIEU2017102}).
 \item segment the cloud into segments, for example by mathematical morphology (see \cite{serna2014detection}) or supervoxel (see \cite{aijazi2013segmentation}), then classify each segment (by hand-made global descriptors \cite{johnson1999using,velizhev2012implicit} or by deep-learning \cite{maturana2015voxnet, qi2016pointnet}).
\end{itemize}
The method used here \cite{roynard2016fast} belongs to the first category. The detailed processing pipeline is:
\begin{itemize}
 \item extraction of the ground by region growing on an elevation map,
 \item segmentation of objects by connectivity of the remaining point cloud,
 \item computation of descriptors on each object (some simple geometric descriptors inspired by \cite{serna2014detection} and some 3D descriptor of the literature as CVFH \cite{rusu2010fast}, GRSD \cite{marton2010general} and ESF \cite{osada2001matching}), 
 \item classification of the objects with a Random Forest.
\end{itemize}

\subsection{Improvements of \cite{roynard2016fast}}

Two improvements are proposed to increase the robustness of this method: first on the segmentation by new extraction of the ground (using better seed for the region growing), then on the classification with new descriptors (to take the context of objects into account).

\subsubsection{Ground Extraction}
In \cite{roynard2016fast}, the seed for region growing is found by computing a histogram in z on the whole cloud, which is not robust in case the road is sloping.
As we know the exact position of the LiDAR sensor with respect to the ground ($2.71\meter$ above ground), we can extract the points that are just below the sensor in a cylinder parameterized by:
\begin{align}
 \sqrt{ (x - x_{origin})^2 + (y - y_{origin})^2 } & \leq 1 \\
 |z_{origin} - z - 2.71| & \leq 0.3
\end{align}
Points lying in this cylinder are then taken as seeds for the region growing.

\subsubsection{Features for Classification}
It was observed that some objects (such as cars) were detected way above the ground. We propose to solve this problem by adding a contextual descriptor which gives the altitude of the object with respect to the ground detected in the previous step.

In a first step we calculate an image of elevation of the ground, for example with a resolution $10\centi\meter \times 10\centi\meter$. Then empty pixels are filled with elevation of the closest non-empty pixel. And the image is smoothed to avoid segmentation artefacts (for example where the ground meets the foot of the buildings).

Then for each object, the barycenter is projected onto this elevation image of the ground, which gives us the elevation of the ground under this object: $z_{ground}$. If $z_{min}$ is the minimum elevation of the object, the descriptor added is: $z_{min} - z_{ground}$.

\subsection{Evaluation: Segmentation}
\begin{center}\centering
 \includegraphics[width=\linewidth]{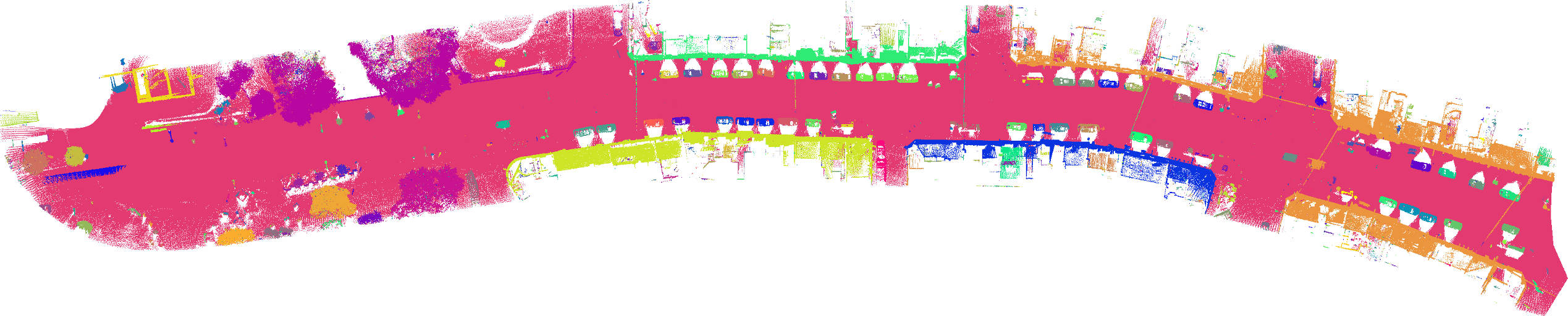}
 \captionof{figure}{Exemple of cloud segmented by our method (each object has a different color).\label{fig:segmEx}}
\end{center}

\begin{figure}[h]\centering
 \includegraphics[width=\linewidth]{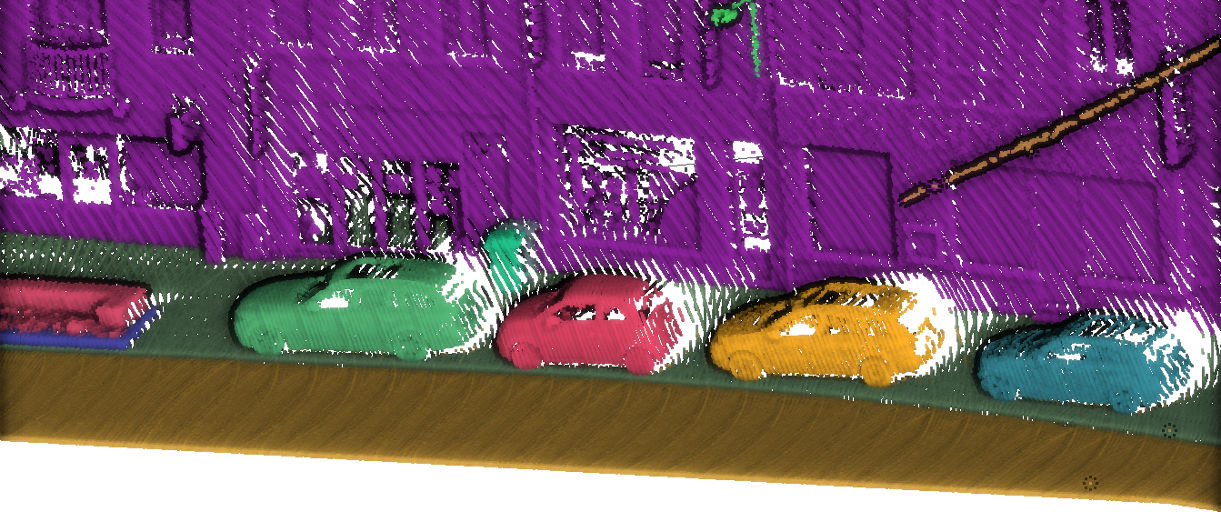}\vspace{0.2cm}
 \includegraphics[width=\linewidth]{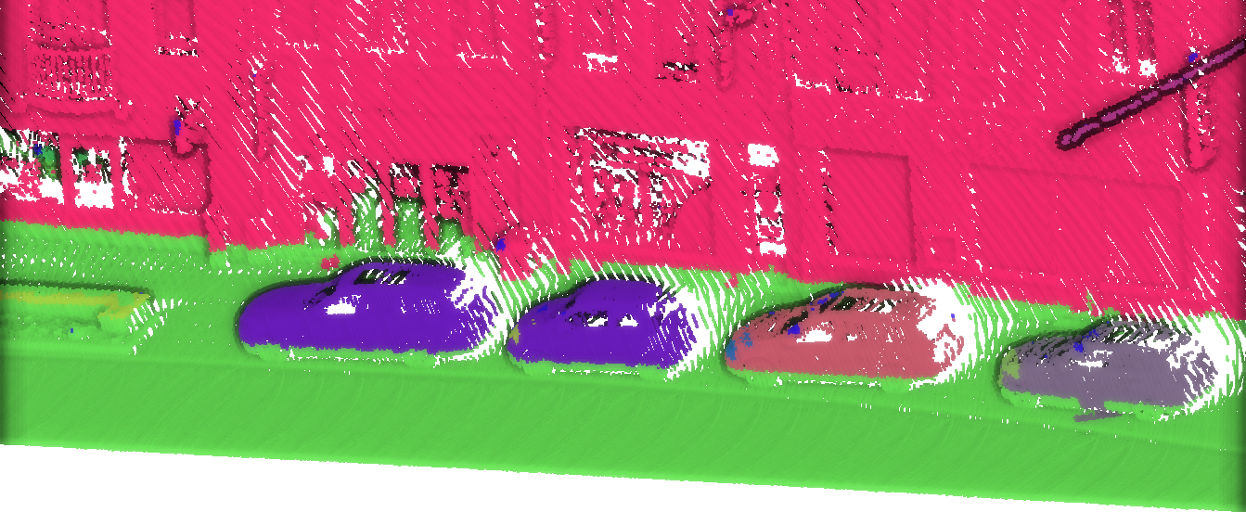}
 \caption{Comparison between clouds segmented automatically by our method (bottom) and by hand (top). Each object has a different color. Two cars too close one from another are segmented as a single object. The bottom part of each object is segmented as part of the ground. A trash can placed against the facade is seen as a part of the facade.\label{fig:segm}}
\end{figure}

Our segmentation method is very basic, indeed it makes very strong a priori on the way to distinguish objects from each other. Two objects are different if they are in different connected component of the point cloud from which the ground has been removed. This explains some problems (see figure \ref{fig:segm}) like two cars too close one from another segmented as a single object, or buildings just linked by a cable.

Our segmentation method is very basic, indeed it makes very strong a priori on the way to distinguish objects from each other. Two objects are different if they are in different connected components of the point cloud from which the ground has been removed. This explains some problems (see figure \ref{fig:segm}) like two cars too close one from another segmented as a single object, or buildings just linked by a cable.


To evaluate detection of objects, we use the same metric as used in iQmulus/TerraMobilita contest \cite{vallet2015terramobilita}.

For an object of the ground truth (represented by the subset $S^{GT}$) and an object resulting from our segmentation method ($S^{SR}$), we estimate that they match if the following conditions are respected:
\begin{equation*}
 \frac{|S^{GT}|}{|S^{GT} \cup S^{SR}|} > m \text{\quad and \quad} \frac{|S^{SR}|}{|S^{GT} \cup S^{SR}|} > m
\end{equation*}

Then detection precision and recall are computed by the following formulas:
\begin{align*} 
 precision(m) &= \frac{\text{number of detected objects matched}}{\text{number of detected objects}} \\
 recall(m) &= \frac{\text{number of detected objects matched}}{\text{number of ground truth objects}} \\
 F1(m) &= \frac{2~ precision(m) \cdot recall(m)}{precision(m) + recall(m)}
\end{align*}

We evaluate our results with $m=0.5$ which is the minimal value that ensures that a Ground Truth object matches at most one object segmented by our method (see table \ref{tab:prec_recall_}).

\begin{center}
 \sf\centering
 \begin{tabular}{cccc}
 \toprule
  Dataset & Precision & Recall & F1 \\
 \midrule
   Lille1  & $70.24 \%$ & $38.55 \%$ & $49.78 \%$\\
   Lille2  & $59.09 \%$ & $31.71 \%$ & $41.27 \%$\\
   Paris   & $54.24 \%$ & $28.46 \%$ & $37.33 \%$\\
 \bottomrule
 \end{tabular}
  \captionof{table}{Precision and Recall of object detection for $m=0.5$.\label{tab:prec_recall_}}
\end{center}

It is believed that methods that learn segmentation will yield much better results.

\subsection{Evaluation: Classification}

\begin{center}\centering
 \includegraphics[width=\linewidth]{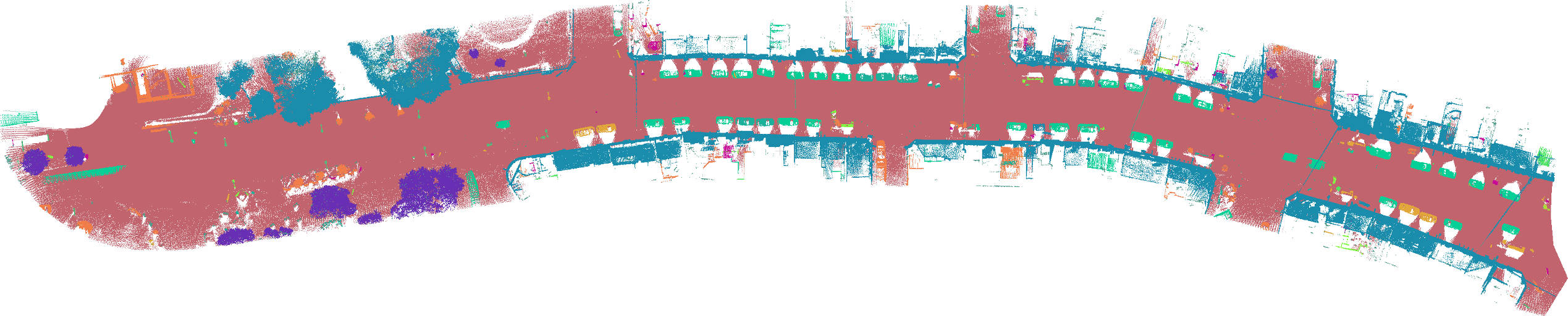}
 \captionof{figure}{Example of cloud classified by our method (each class has a different color).\label{fig:classEx}}
\end{center}

In this section we only evaluate the classification method assuming good segmentation. To do this, we take the set of objects of the dataset that are randomly divided into a training set ($80\%$) and a test set ($20\%$).
We use only a few coarser classes than described in table \ref{tab:nb_samp} to evaluate our classification algorithm, see table \ref{tab:coarse_classes} for a distribution of samples per class. In addition, we add a \texttt{coarse\_classes.xml} file to the dataset that adds a \texttt{coarse} field to each class.

\begin{center}
 \includegraphics[width=\linewidth]{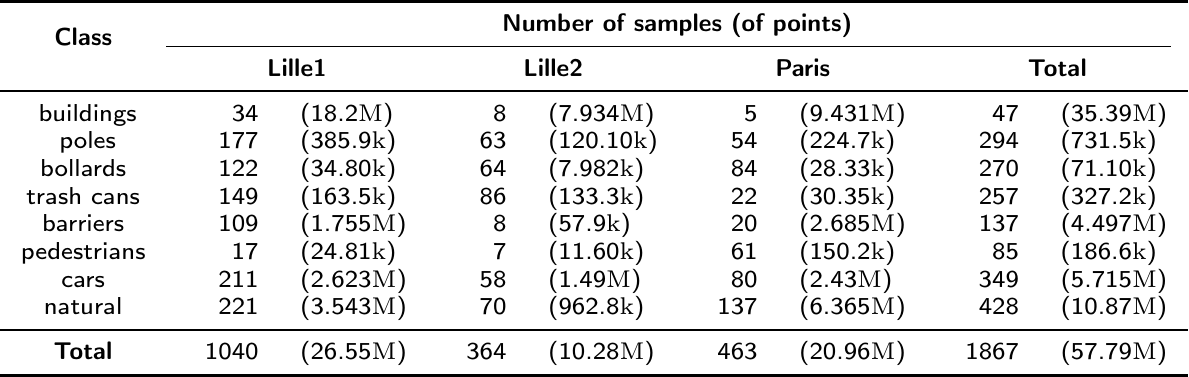}
 \captionof{table}{Number of samples/points for each coarse class used for classification evaluation.\label{tab:coarse_classes}}
\end{center}

Even with these coarse classes, there are a few samples in some of them. Then precision and recall numbers in table \ref{tab:prec_recall_desc_classif} should be taken with caution. Metrics used to evaluate performance are the following:
\begin{align} \label{equ:indicators}
 P   &= \frac{TP}{TP + FP}\nonumber \\
 R   &= \frac{TP}{TP + FN} \\
 F1  &= \frac{2TP}{2TP + FP + FN}\nonumber \\
 MCC &= \frac{TP\cdot TN - FP\cdot FN}{\sqrt{(TP + FP)(TP + FN)(TN + FP)(TN + FN)}}\nonumber
\end{align}
Where $P$, $R$, $F1$ and $MCC$ represent respectively Precision, Recall, F1-score and Matthews correlation coefficient. And $TP$, $TN$, $FP$ and $FN$ are respectively the number of True-Positives, True-Negatives, False-Positives and False-Negatives.

 Moreover, it can be noted that the best results are obtained with the combination of descriptors: Geometric and GRSD, which are the descriptors composed of the least number of variables. 
This can be explained by the few samples of the dataset and therefore adding a large number of features does not provide more relevant information.
Then this dataset is more appropriate for the evaluation of per-point classification methods.
  
\begin{center}
 \includegraphics[width=\linewidth]{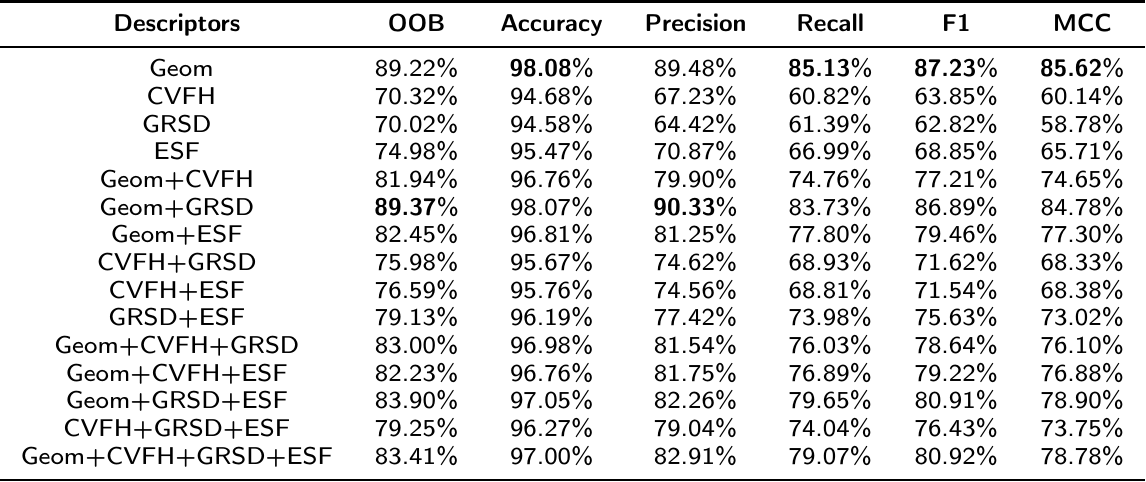}
 \captionof{table}{Classification performance for each combination of descriptors, these metrics are averaged over all classes (the OOB score is given by Random-Forest during training).\label{tab:prec_recall_desc_classif}}
\end{center}

It can be concluded that it is not necessary to calculate all the descriptors to obtain the best classification results. It is possible to gain in computation time by calculating only the geometric descriptors and GRSD (see table \ref{tab:tps_descr} for precise gains). And for applications where time is critical, we can even calculate only the geometric descriptors (which also avoids having to calculate the normals).

\begin{center}
  \centering
  \begin{tabular}{cccc}
  \toprule
     \textbf{Descriptors} & \textbf{Proportion} & \textbf{Mean Time per object (ms)} \\
  \midrule  
   Geom		    	& $3.22\%$	& $0.9$  \\
  \midrule  
   CVFH				& $44.92\%$	& $11.9$ \\
  \midrule  
   GRSD				& $12.04\%$	& $3.2$  \\
  \midrule  
   ESF				& $39.82\%$	& $10.6$ \\
  \midrule  
   \textbf{Total}	& $100\%$	& $26.6$ \\
  \midrule \midrule
   \textbf{Normals}  & & $47.9$\\
  \bottomrule 
 \end{tabular}
  \captionof{table}{Mean computational time for calculating descriptors on segmented objects. Time to compute normal vectors is added for comparison.\label{tab:tps_descr}}
\end{center}

\section{Conclusion}

We presented a dataset of urban 3D point cloud for automatic segmentation and classification. This dataset contains 140 million points on $2\kilo\meter$ in two different cities. The objects were segmented by hand and a class was associated with each one among 50 classes.

We hope that this dataset will help to train and evaluate methods as deep-learning, which are very demanding in terms of quantity of points.

In addition, we have tested a first method of segmentation and automatic classification from \cite{roynard2016fast} to which we have made some improvements for robustness.

\bibliographystyle{alpha}
\bibliography{dataset_Roynard_2017}

\end{document}